\documentclass[]{interact}

\usepackage{epstopdf}  
\usepackage{subfigure} 
\usepackage{float}
\usepackage{amsmath, amssymb}
\usepackage{graphicx}
\usepackage{algorithm2e}
\usepackage{hyperref}
\usepackage{natbib}
\bibpunct[, ]{(}{)}{;}{a}{}{,}
\usepackage{lmodern}
\renewcommand\bibfont{\fontsize{10}{12}\selectfont}
\usepackage{tabularx}

\theoremstyle{plain}

\theoremstyle{definition}

\theoremstyle{remark}

\begin{document}

\articletype{RESEARCH ARTICLE}

\title{A GAN-Enhanced Deep Learning Framework for Rooftop Detection from Historical Aerial Imagery}

\author{
\name{Pengyu Chen\textsuperscript{a},  Sicheng Wang\textsuperscript{a}\thanks{CONTACT Sicheng Wang. Email: SICHENGW@mailbox.sc.edu}, Cuizhen Wang\textsuperscript{a}, Senrong Wang\textsuperscript{b}, Beiao Huang\textsuperscript{c}, Lu Huang\textsuperscript{b}, and Zhe Zang\textsuperscript{d}}
\affil{\textsuperscript{a}Department of Geography, University of South Carolina, Columbia, SC, USA;\\
\textsuperscript{b}School of Computer Science and Engineering, Wuhan University of Technology, Wuhan, China;\\
\textsuperscript{c}School of Remote Sensing and Information Engineering, Wuhan University, Wuhan, China;\\
\textsuperscript{d}College of Marine Geosciences, Ocean University of China, Qingdao, China}
}

\maketitle

\begin{abstract}
Precise detection of rooftops from historical aerial imagery is essential for analyzing long-term urban development and human settlement patterns. Nonetheless, black-and-white analog photographs present considerable challenges for modern object detection frameworks due to their limited spatial resolution, absence of color information, and archival degradation. To address these challenges, this research introduces a two-stage image enhancement pipeline based on Generative Adversarial Networks (GANs): image colorization utilizing DeOldify, followed by super-resolution enhancement with Real-ESRGAN. The enhanced images were subsequently employed to train and evaluate rooftop detection models, including Faster R-CNN, DETReg, and YOLOv11n. The results demonstrate that the combination of colorization with super-resolution significantly enhances detection performance, with YOLOv11n achieving a mean Average Precision (mAP) exceeding 85\%. This signifies an enhancement of approximately 40\% over the original black-and-white images and 20\% over images enhanced solely through colorization. The proposed method effectively bridges the gap between archival imagery and contemporary deep learning techniques, facilitating more reliable extraction of building footprints from historical aerial photographs. Code and resources for reproducing our results are publicly available at \href{https://github.com/Pengyu-gis/Historical-Aerial-Photos}{github.com/Pengyu-gis/Historical-Aerial-Photos}.
\end{abstract}


\newpage
\section{Introduction}
Remote sensing has been a cornerstone for understanding and managing the built environment, enabling applications such as urban planning, infrastructure monitoring, and disaster response \citep{wellmann_remote_2020, tosti_integration_2021, pi_convolutional_2020}. Among these, rooftop detection plays a vital role in extracting building footprints, assessing structural integrity, and informing policy decisions related to land use and hazard mitigation \citep{yuan_building_2025, bauchet_rooftops_2021, ding_land-useland-cover_2013, norman_review_2019}. 

The United States started to collect aerial photographs since the early 20th century.  These historical aerial images provide rich information about past built environments. For example, the United States Geological Survey (USGS) began to acquire aerial photographs in 1937 \citep{us_geological_survey_us_1973}. These archival images are invaluable for tracking the urban development history of many cities and regions, offering insights into growth patterns, spatial structure transformation, and long-term infrastructural changes \citep{farella_eurosdr_2022}.

Recent advances and applications of aerial image enhancement techniques, such as colorization and super-resolution, have seen remarkable progress with deep learning, demonstrating promise for overcoming the limitations of historical imagery. Colorization methods, especially Generative Adversarial Networks (GANs) and Stable Diffusion, can significantly enhance visual interpretability by introducing plausible color channels to grayscale images \citep{goodfellow_generative_2014, rombach_high-resolution_2022}. For instance, DeOldify, a widely adopted GAN-based approach, can effectively reduce artifacts while preserving structural details in archival images \citep{antic_janticdeoldify_2025}. Additionally, super-resolution techniques like Real-ESRGAN address low-resolution constraints by reconstructing high-resolution details, thereby improving object detection by refining edges and textures \citep{wang_real-esrgan_2021, haut_new_2018, dong_rrsgan_2022, zhang_efficient_2024}.

Various rooftop detection architectures have emerged and evolved, including Convolutional Neural Networks (CNNs) and Transformers, each tailored to specific detection needs. CNN-based models, such as Faster R-CNN, Mask R-CNN, and YOLO variants, have been demonstrated effective. For example, Faster R-CNN has accurately delineated rooftops in medium-resolution aerial images, whereas YOLO models have provided rapid, reliable detection suitable for large-scale applications \citep{ren_faster_2017, he_mask_2018, park_detecting_2024, avudaiamal_yolo_2024, khanam_yolov11_2024}. Transformer-based models, including DETR and Vision Transformer (ViT), capture both local and global contexts, beneficial for complex urban rooftop detection scenarios \citep{carion_end--end_2020, bar_detreg_2023, dosovitskiy_image_2021, yuan_building_2025}.

\begin{figure}[H]
    \centering
    \includegraphics[width=1\linewidth]{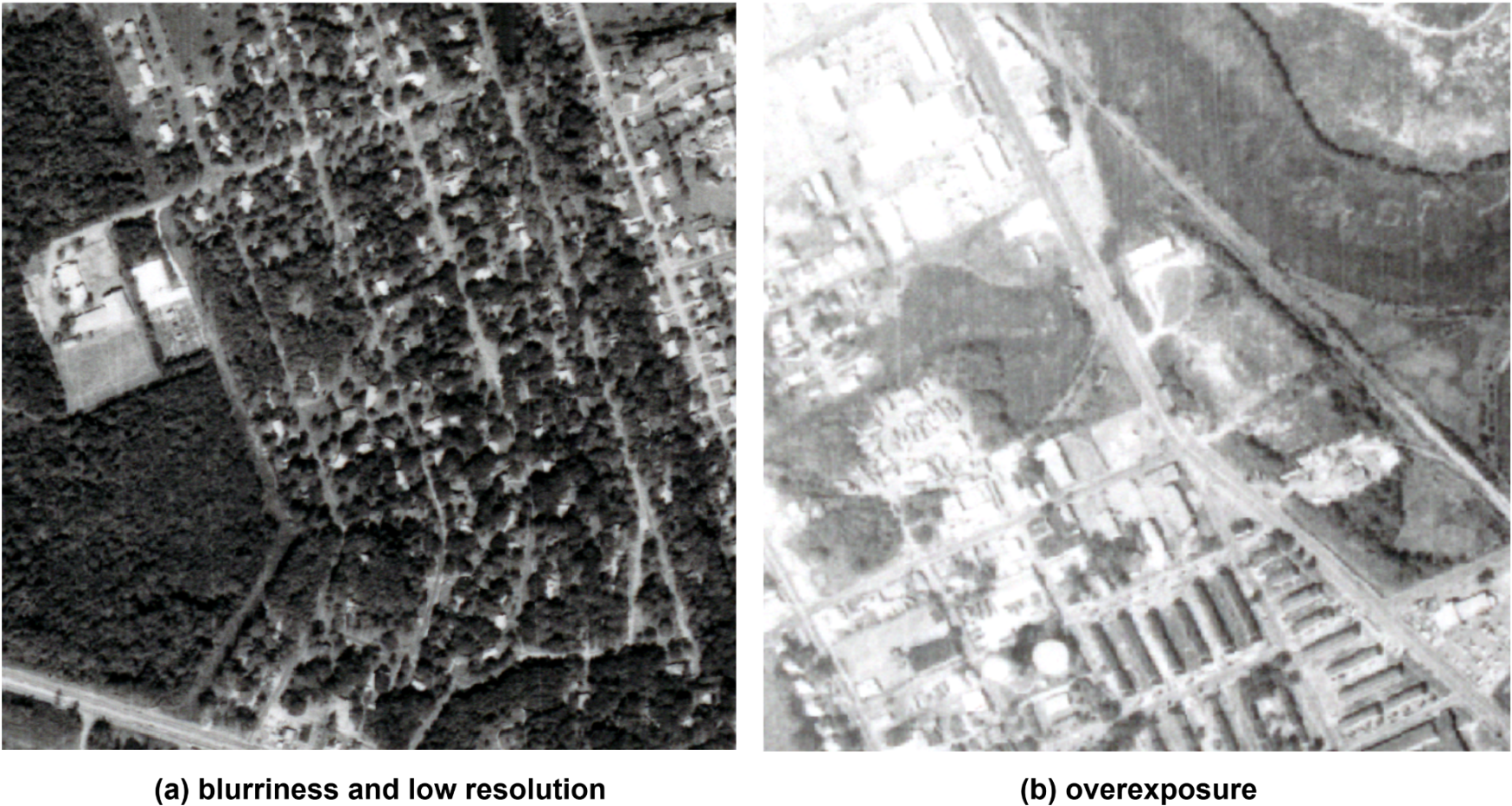}
    \caption{Limitations of Historical Black-and-White Aerial Images.\\
    Source: \citet{usc_library_south_2024}.}
    \label{fig:limits}
\end{figure}

Although existing rooftop detection methodologies have shown effectiveness, a significant gap remains in the exploration of historical aerial imagery, which presents unique challenges due to limited quality and low resolution \cite{fertel_vegetation_2023}. Furthermore, few models are optimized explicitly for grayscale images, which are prevalent in archival datasets. As illustrated in Figure~\ref{fig:limits}, two common issues are exemplified: the left image suffers from blurriness and low resolution, impairing object identification and feature extraction; the right image exhibits overexposed regions that obscure critical structural details. These limitations often result in discrepancies between detected and actual building outlines. In response to the lack of a standardized, scalable workflow for processing such images, this study introduces a two-stage enhancement pipeline. The proposed approach leverages GAN-based frameworks to sequentially apply image colorization and super-resolution, aiming to improve the fidelity of feature extraction from degraded historical imagery.

The rationale behind this two-stage process lies in aligning historical imagery with the data standards expected by modern deep learning models, which are typically trained on high-resolution, RGB-colored imagery. Colorization serves as the first step to reconstruct the chromatic information lost in early aerial imagery, thus enabling the models to exploit color-related features such as material reflectance or shadow hue, which are critical for distinguishing rooftops from other urban surfaces \citep{liu_single_2018, fu_satellite_2024, farella_colorizing_2022}.

Super-resolution is applied after colorization because color information provides additional spatial priors that improve the quality of upscaling. Previous studies have shown that applying super-resolution directly to grayscale images often leads to artifacts or poor reconstruction of semantic features, whereas colored inputs yield sharper and more accurate high-resolution outputs \citep{liang_swinir_2021, konovalov_study_2024, gao_high-quality_2025}. This sequence thus maximizes the fidelity of restored features crucial for object detection tasks \citep{feng_deep_2022}.

This enhancement pipeline transforms black-and-white historical aerial images into rich, high-quality data, enabling more accurate rooftop detection and paving the way for long-term geospatial analysis in urban planning, disaster resilience, and architectural history.

\section{Study Area and Data}

\subsection{Study area: Charleston, South Carolina}

Charleston, the largest city in South Carolina, is a historic port city founded in 1670 on the southeastern U.S. coast, crucial in colonial and post-colonial times. Its well-preserved architecture, including Colonial, Georgian, Federal, Greek Revival, Italianate, and Victorian styles, makes it ideal for studying urban transformations and rooftop detection from historical imagery\citep{baco_historic_2009}. Charleston's densification, expansion, and infrastructure evolution provide insights into urban morphology and spatial changes, serving as a benchmark for historical urban studies \citep{baco_historic_2009, rybczynski_charleston_2019, hanchett_sorting_2020}.

\subsection{Historical aerial imagery dataset in 1979}

\begin{figure}[H]
    \centering
    \includegraphics[width=1\linewidth]{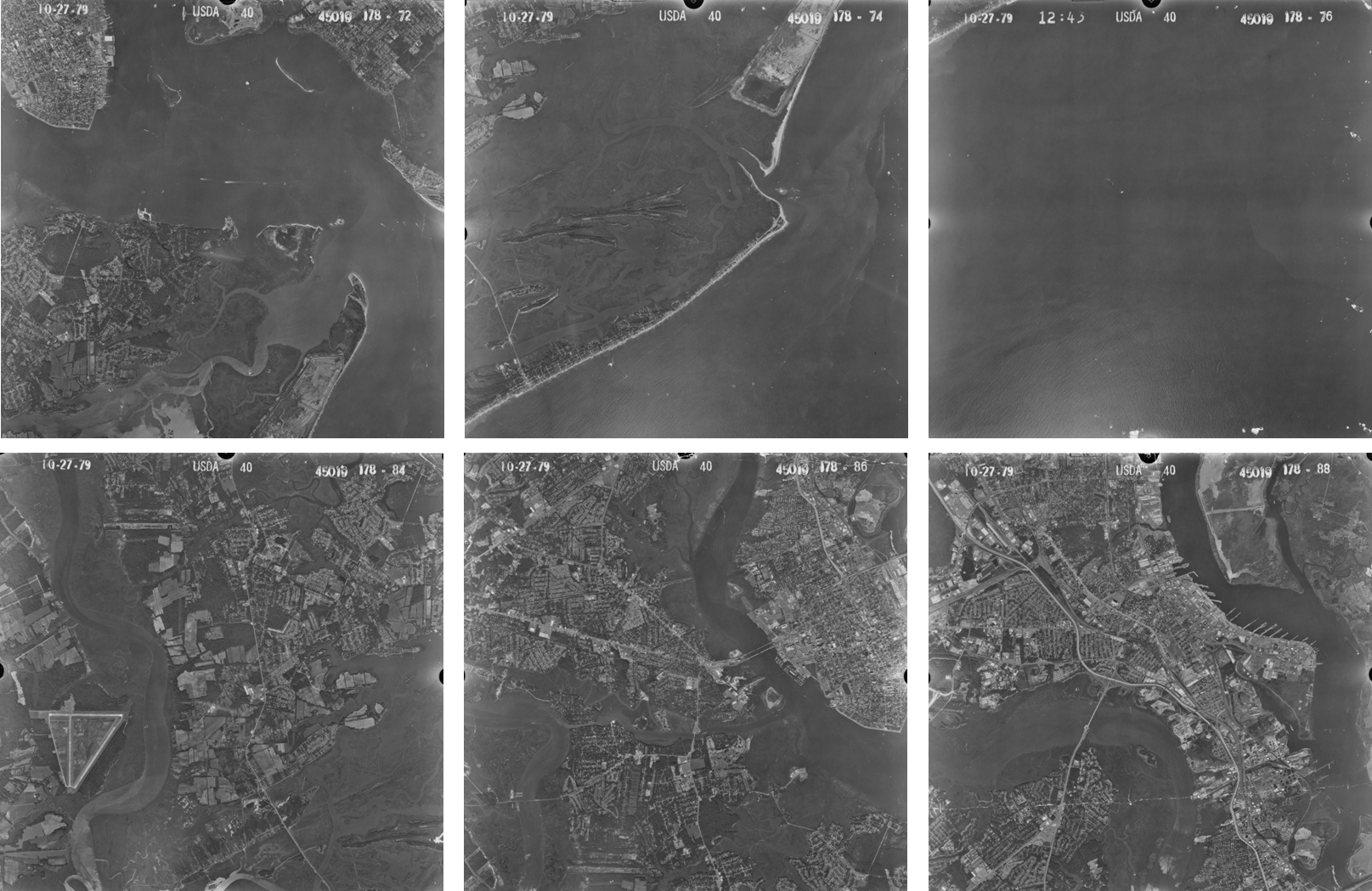}
    \caption{Samples of historical black-and-white aerial photographs captured over Charleston, South Carolina in 1979.\\
    Source: \citet{usc_library_south_2024}.}
    \label{fig:1979images}
\end{figure}

The primary dataset utilized in this study consists of historical black-and-white aerial photographs taken in 1979 through surveys conducted by the United States Department of Agriculture (USDA). These images were sourced from the digitized archives curated by the Thomas Cooper Library at the University of South Carolina (USC) \citep{usc_library_south_2024}. The photographs were digitized and geo-referenced by the USC librarians, facilitating their spatial alignment with contemporary datasets. The imagery from 1979 offers an invaluable portrayal of the built environment in downtown Charleston during a significant phase of urban development and sprawl. Although the spatial resolution is inferior to that of more recent satellite imagery, it is adequate for detecting building footprints when suitably enhanced. Figure~\ref{fig:1979images} displays selected samples of the historical aerial imagery employed in this study.

\subsection{Ground truth and reference data}

To assess the accuracy of rooftop detection, the Microsoft GlobalMLBuildingFootprints dataset \citep{heris_rasterized_2020} was employed as the ground truth. This dataset encompasses a comprehensive and high-resolution assemblage of building footprints derived from recent satellite imagery via machine learning techniques. Covering the entirety of the United States, it provides high geometric precision and semantic consistency, thus serving as a reliable reference for evaluating the detection quality of rooftops in historical imagery. Given the temporal gap between the 1979 aerial photographs and the current building footprints, a subset of the reference data was manually revised to ensure accurate alignment in regions where the urban form has undergone significant changes.

\section{Methodology}
\begin{figure}[H]
    \centering
    \includegraphics[width=1\linewidth]{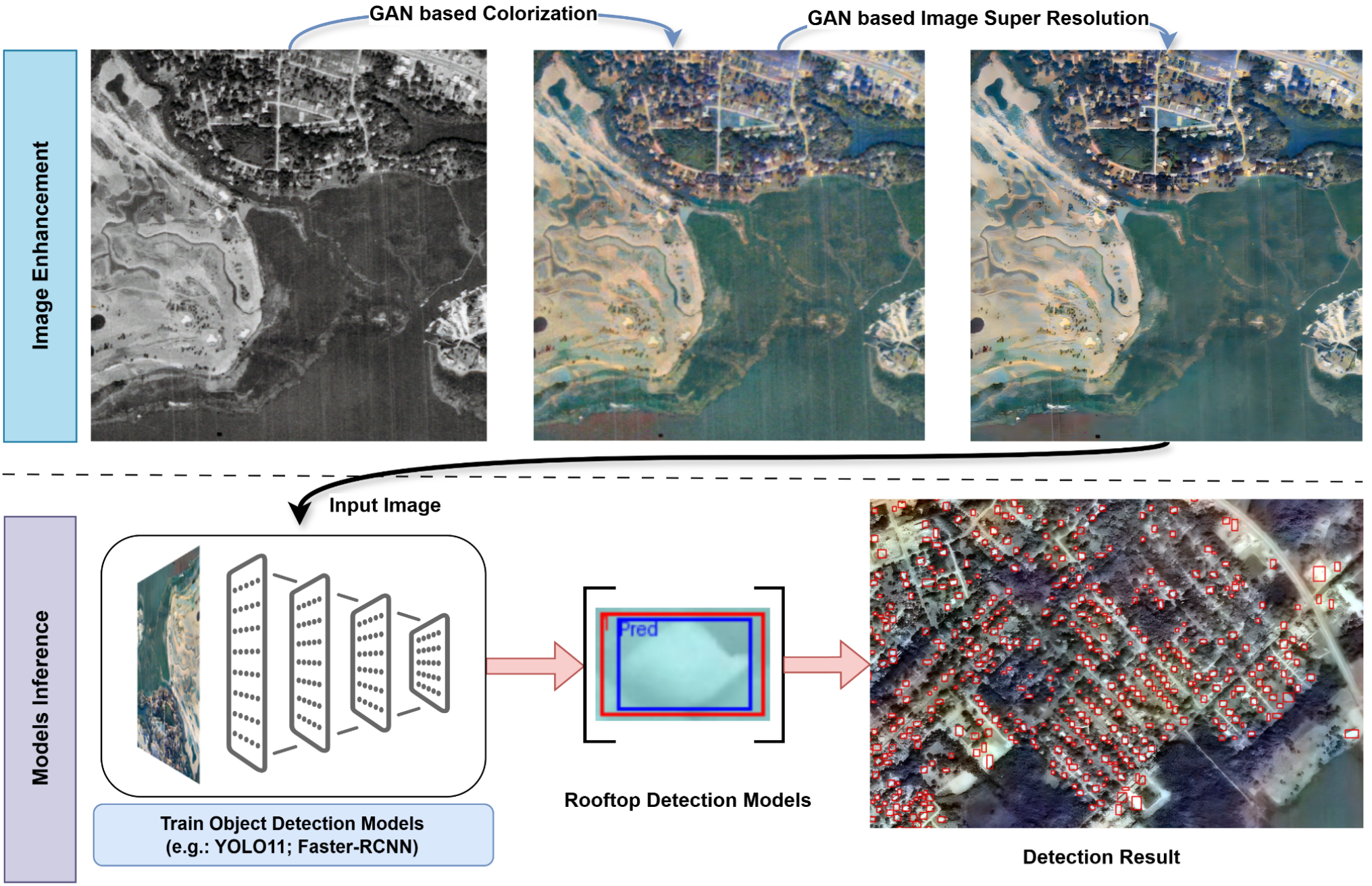}
    \caption{Overall workflow for rooftop detection in historical aerial imagery. The pipeline starts with image enhancement (colorization and super-resolution), followed by model training, and concludes with bounding box outputs.\\
Source: Graph illustrated by authors}
    \label{fig:framework}
\end{figure}

This section outlines a complete workflow for image preprocessing and rooftop detection based on historical aerial imagery, focusing on image quality enhancement to improve detection accuracy. As illustrated in the flow diagram in Figure~\ref{fig:framework}, our process involves two steps of image enhancement (colorization and super-resolution), subsequent model training, and the eventual production of bounding box outputs for detected rooftops.
\label{sec:methodology}

\subsection{Image enhancement approaches}
To address the challenges of historical aerial imagery, we applied two GAN-based image enhancement techniques: (1) colorization to introduce plausible chromatic details and (2) super-resolution imaging to improve spatial resolution and clarity.

\subsubsection{Image colorization using Deoldify}
The objective of colorization is to (1) enhance the visual contrast and interpretability of historical black-and-white aerial images and (2) generate images compatible with pretrained object detection models, typically trained on RGB imagery. We employed DeOldify, a GAN architecture optimized for historical image restoration \citep{antic_janticdeoldify_2025, goodfellow_generative_2014}.

DeOldify's architecture combines a deep generator network with a relativistic discriminator. The generator employs a U-Net structure built on a ResNet-34 backbone,  augmented by self-attention mechanisms to better handle spatial context \citep{antic_janticdeoldify_2025}. This enables consistent color hypotheses for large urban features like contiguous rooftops and building footprints. Rather than operating in RGB space, the framework processes images in the CIELAB color space, preserving the original luminance channel while predicting chrominance components. Equation (1) presents the generator \( G \) predicts the chrominance components \( I_{ab} \) as follows:
\begin{equation}
    \hat{I}_{ab} = G(I_L).
\end{equation}
where \( I_L \) represents the grayscale (luminance) image. The final colored image is obtained by merging the preserved luminance with the predicted chrominance, as expressed by Equation (2):
\begin{equation}
    I_{\text{colored}} = \left[ I_L, \hat{I}_{ab} \right].
\end{equation}

The adversarial component uses a PatchGAN discriminator \( D \) that evaluates local image regions rather than the full image, enforcing high-frequency realism in architectural elements such as roof edges, texture patterns, and material boundaries. The GAN loss can be formulated as Equation (3):
\begin{equation}
    \mathcal{L}_{\text{GAN}}(G, D) = \mathbb{E}_{I_{\text{color}}} \left[ \log D(I_{\text{color}}) \right] + \mathbb{E}_{I_L} \left[ \log \left(1 - D\left(\left[ I_L, G(I_L) \right]\right) \right) \right],
\end{equation}
where \( I_{\text{color}} \) represents a true color image. In addition, a pixel-wise reconstruction loss (e.g., L1 loss) is employed in the process of Equation (4):
\begin{equation}
    \mathcal{L}_{L1}(G) = \mathbb{E}_{I_{\text{color}}, I_L}\left[ \| I_{\text{color}} - \left[ I_L, G(I_L) \right] \|_1 \right].
\end{equation}

Self-attention layers are integrated into the generator. A typical self-attention operation is given by Equation (5):
\begin{equation}
    \text{Attention}(Q, K, V) = \text{softmax}\left(\frac{QK^\top}{\sqrt{d_k}}\right)V,
\end{equation}
where \( Q \), \( K \), and \( V \) are the query, key, and value feature maps, and \( d_k \) is the dimensionality of the keys. This mechanism allows the model to capture global contextual information, which is critical for accurate colorization of extended urban features.

\begin{figure}[H]
\centering
\includegraphics[width=\linewidth]{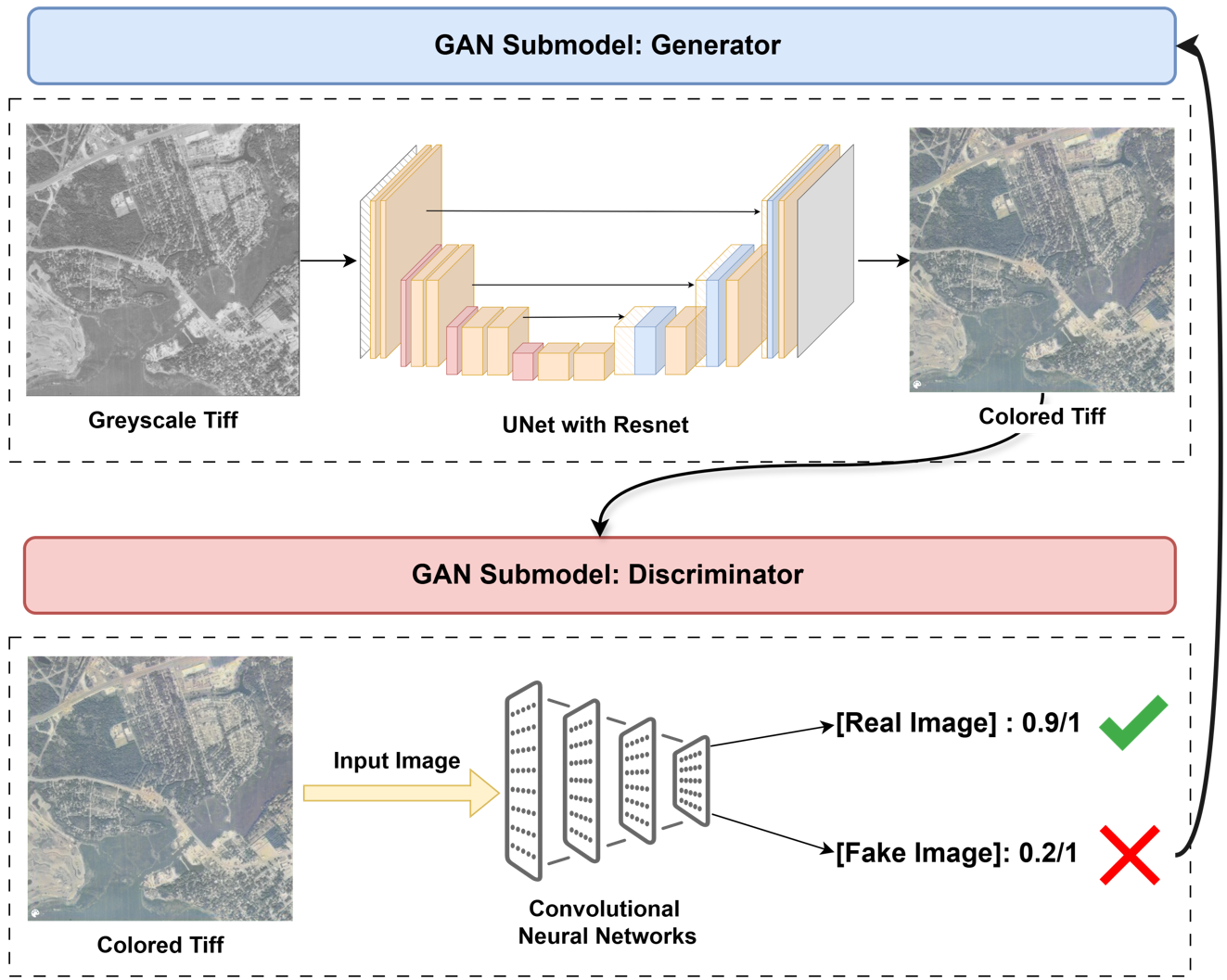}
\caption{DeOldify architecture for aerial image colorization. The generator (top pathway) processes luminance values through residual blocks and self-attention layers to predict chrominance, while the PatchGAN discriminator (bottom pathway) evaluates local realism. Structural details are preserved through direct retention of the input luminance channel.\\
Source: Graph illustrated by authors.}
\label{fig:colorization_process}
\end{figure}

Figure~\ref{fig:colorization_process} illustrates the workflow of colorization: the generator ingests the grayscale input and iteratively predicts chrominance values through a cascade of residual blocks and attention gates. Concurrently, the discriminator analyzes overlapping image patches to assess local color realism. The final output merges the original luminance with synthesized chrominance, producing a colored image where roof geometries remain aligned with the source data while gaining plausible hue/saturation characteristics.

\subsubsection{Super-resolution imaging with Real-ESRGAN}

To further improve the clarity of building details, we upscaled the colored aerial images using Real-ESRGAN, a GAN-based super-resolution model \citep{wang_real-esrgan_2021}. Real-ESRGAN is specifically designed to handle real-world image noise and artifacts, making it particularly suitable for historical aerial imagery. By increasing the spatial resolution of the input images, we enhanced the visibility of critical rooftop features, thereby improving both precision and recall during the object detection training phase.

Real-ESRGAN follows a generative adversarial framework, which can be expressed as Equation (6):
\begin{equation}
    I_{SR} = G_{SR}(I_{LR}).
\end{equation}
where \( I_{LR} \) denotes the low-resolution (colored) image, and \( I_{SR} \) represents the super-resolved image produced by the generator \( G_{SR} \).
The generator applies a series of residual blocks and upsampling layers to produce high-resolution outputs, while the discriminator \( D_{SR} \) evaluates the realism of these upscaled images. The GAN loss for super-resolution is given by Equation (7):
\begin{equation}
    \mathcal{L}_{\text{GAN}}^{SR}(G_{SR}, D_{SR}) = \mathbb{E}_{I_{HR}} \left[ \log D_{SR}(I_{HR}) \right] + \mathbb{E}_{I_{LR}} \left[ \log \left(1 - D_{SR}\left(G_{SR}(I_{LR})\right)\right) \right],
\end{equation}
where \( I_{HR} \) represents the ground truth high-resolution image. A reconstruction loss is also used to enforce pixel-level accuracy as expressed in Euqation (8):
\begin{equation}
    \mathcal{L}_{\text{rec}}(G_{SR}) = \| I_{HR} - I_{SR} \|_1.
\end{equation}

\begin{figure}[H]
    \centering
    \includegraphics[width=\linewidth]{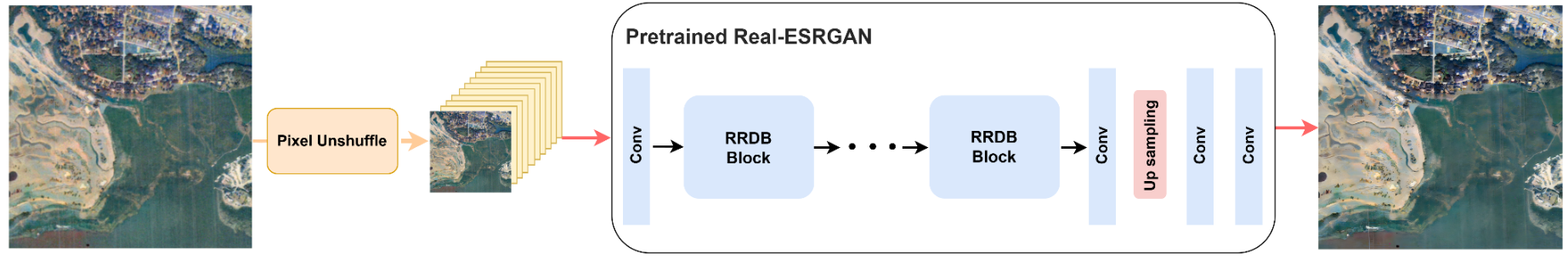}
    \caption{How Real-ESRGAN work for super-resolution.\\
Source: Graph illustrated by authors.}
    \label{fig:upscale_workflow}
\end{figure}

Figure~\ref{fig:upscale_workflow}  depicts the working mechanism of Real-ESRGAN. The adversarial training strategy encourages the generator to produce images with sharper edges and more faithful textures, ensuring that rooftop details are preserved for subsequent detection tasks. In our experiments, applying Real-ESRGAN led to higher mean average precision (mAP) and recall rates when training the rooftop detection model.

\subsection{Object detection model training}
\label{subsec:object_detection}

\begin{figure}[H]
    \centering
    \includegraphics[width=1\linewidth]{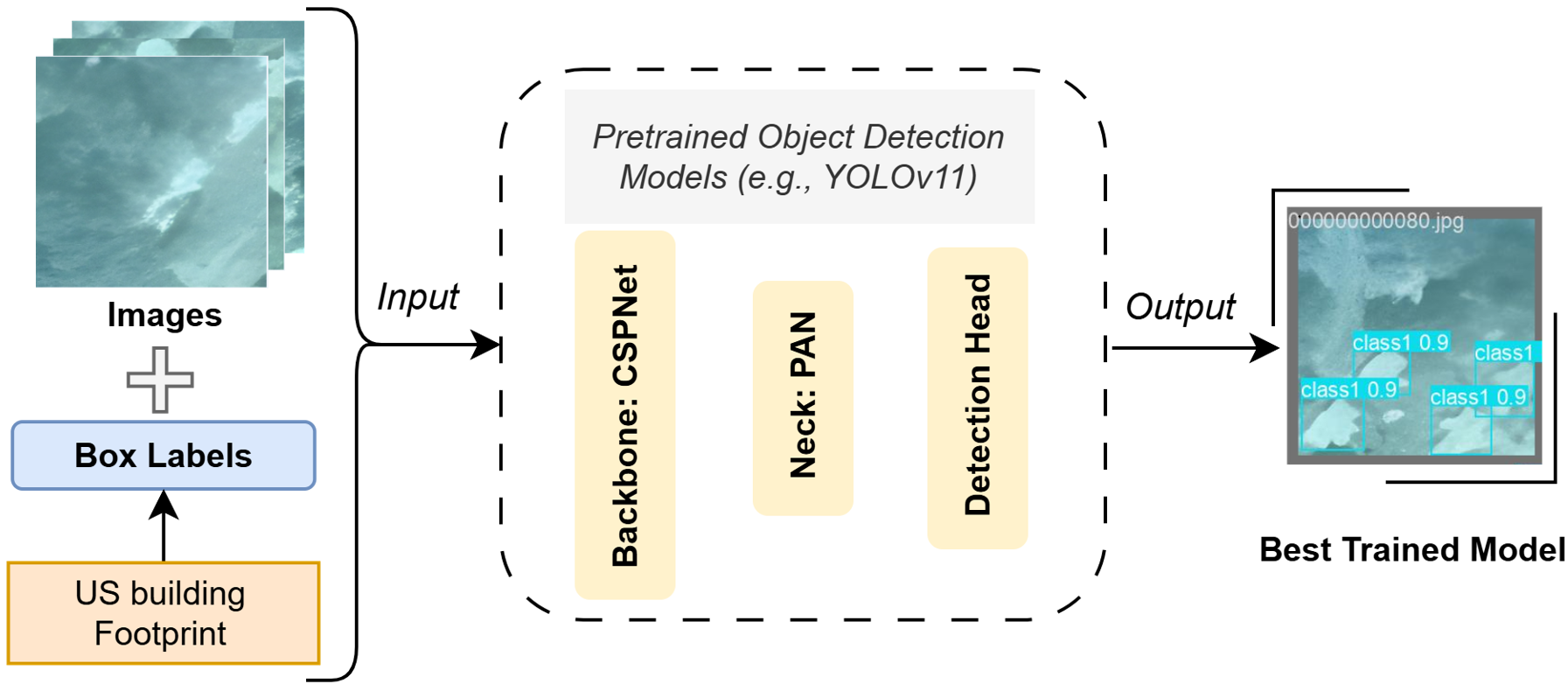}
    \caption{Overview of the object detection model training process. The input consists of aerial images and bounding box annotations derived from Microsoft's US Building Footprints dataset. The model follows a typical COCO-style architecture with a feature extraction backbone, a PAN-based neck for feature fusion, and a detection head that outputs bounding boxes with confidence scores.\\
Source: Graph illustrated by authors.}
    \label{fig:Yolo11}
\end{figure}

In this study, we fine-tuned three object detection models: YOLOv11, Faster R-CNN, and DETR \citep{ren_faster_2017, bar_detreg_2023, khanam_yolov11_2024}. The training process follows a standardized pipeline, with YOLOv11n used as an illustrative example. The training process is shown in Figure~\ref{fig:Yolo11}.

For our YOLO-based framework, let each predicted bounding box be represented by Equation (9):
\begin{equation}
    \hat{b} = (\hat{x}, \hat{y}, \hat{w}, \hat{h}, \hat{c}, \hat{p}),   
\end{equation}
where \((\hat{x}, \hat{y})\) is the center coordinate, \(\hat{w}\) and \(\hat{h}\) are the width and height, \(\hat{c}\) is the object confidence, and \(\hat{p}\) is the predicted class probability distribution.
The overall training loss is a combination of localization, confidence, and classification losses, as expressed by Equation (10):
\begin{equation}
    \mathcal{L} = \lambda_{box} \mathcal{L}_{box} + \lambda_{obj} \mathcal{L}_{obj} + \lambda_{cls} \mathcal{L}_{cls},
\end{equation}
where the hyperparameters \(\lambda_{box}\), \(\lambda_{obj}\), and \(\lambda_{cls}\) balance the contributions of each component.
The localization loss \(\mathcal{L}_{box}\) is computed using a bounding box regression loss (e.g., Complete Intersection over Union (CIoU) loss) specified in Equation (11):
\begin{equation}
    \mathcal{L}_{box} = \sum_{i=1}^{N} \mathbb{I}_i^{obj} \cdot \text{CIoU}(b_i, \hat{b}_i),
\end{equation}
where \(\mathbb{I}_i^{obj}\) is an indicator function that equals 1 if object \(i\) is present, and \(b_i\) is the ground truth bounding box.
The confidence loss \(\mathcal{L}_{obj}\) is modeled using a binary cross-entropy loss as formulated in Equation (12):
\begin{equation}
    \mathcal{L}_{obj} = -\sum_{i=1}^{N} \left[ \mathbb{I}_i^{obj} \log \hat{c}_i + (1 - \mathbb{I}_i^{obj}) \log (1 - \hat{c}_i) \right],
\end{equation}

This ensures the model learns to differentiate between object and background regions.
The classification loss \(\mathcal{L}_{cls}\) is defined using the categorical cross-entropy formulated as Equation (13):
\begin{equation}
    \mathcal{L}_{cls} = -\sum_{i=1}^{N} \mathbb{I}_i^{obj} \sum_{c=1}^{C} y_{i,c} \log \hat{p}_{i,c},
\end{equation}
where \(C\) is the number of classes, \(y_{i,c}\) is the one-hot encoded ground truth, and \(\hat{p}_{i,c}\) is the predicted probability for class \(c\).

In our model implementation, the Cross Stage Partial Network (CSPNet) is used as the backbone for efficient feature extraction. At the same time, a Path Aggregation Network (PAN) serves as the neck to enhance multi-scale feature fusion. As mathematically summarized above, the detection head then processes these features to output the final bounding box predictions.

\subsubsection{Dataset preparation}  
To generate suitable training data, we converted vector building footprints provided by Microsoft's USBuildingFootprints dataset into minimum bounding rectangles (MBRs). Bounding boxes were manually refined where necessary to align precisely with rooftop edges visible in historical imagery. The finalized annotations were formatted using the COCO standard \citep{lin_microsoft_2015}, facilitating seamless integration with common deep-learning frameworks.

\subsubsection{YOLOv11 model training procedure}
We initialized training with a pretrained YOLOv11n model, leveraging transfer learning to adapt the detector to historical aerial imagery. The training configurations and parameters are shown in Table~\ref{tab:training_config}.

\begin{table}[htbp]
    \centering
    \caption{Training Configuration for Object Detection Models}
    \label{tab:training_config}
    \begin{tabular}{l  p{8cm}}
        \hline
        \textbf{Parameter} & \textbf{Value} \\
        \hline
        Optimizer & Adam \\
        Initial Learning Rate & \(1 \times 10^{-3}\) \\
        Batch Size & 16 images per batch \\
        Training Epochs & 100 (with early stopping) \\
        Data Augmentation & Rotation, flipping, scaling, brightness/contrast normalization \\
        Loss Function & \(\mathcal{L} = \mathcal{L}_{\text{cls}} + \mathcal{L}_{\text{box}} + \mathcal{L}_{\text{obj}}\) \\
        Annotation Format & COCO \\
        Evaluation Metrics & mAP, Precision, Recall, F1-score \\
        \hline
    \end{tabular}
\end{table}

This standardized approach ensured consistency and fairness across comparisons among YOLOv11, Faster R-CNN, and DETR, allowing reliable evaluation of each model's performance in rooftop detection tasks using historical aerial imagery.

\section{Results}
In this section, we present our results on image enhancement through colorization and image upscaling, as well as the deep learning model training for rooftop detection based on the 1979 Charleston aerial photographs. Our experiments demonstrate that both colorization and super-resolution are not only helpful but also essential for enhancing image quality, which in turn improves the performance of our object detection models.

\subsection{Black-white aerial image colorization results}

\begin{figure}[H]
    \centering
    \includegraphics[width=1\linewidth]{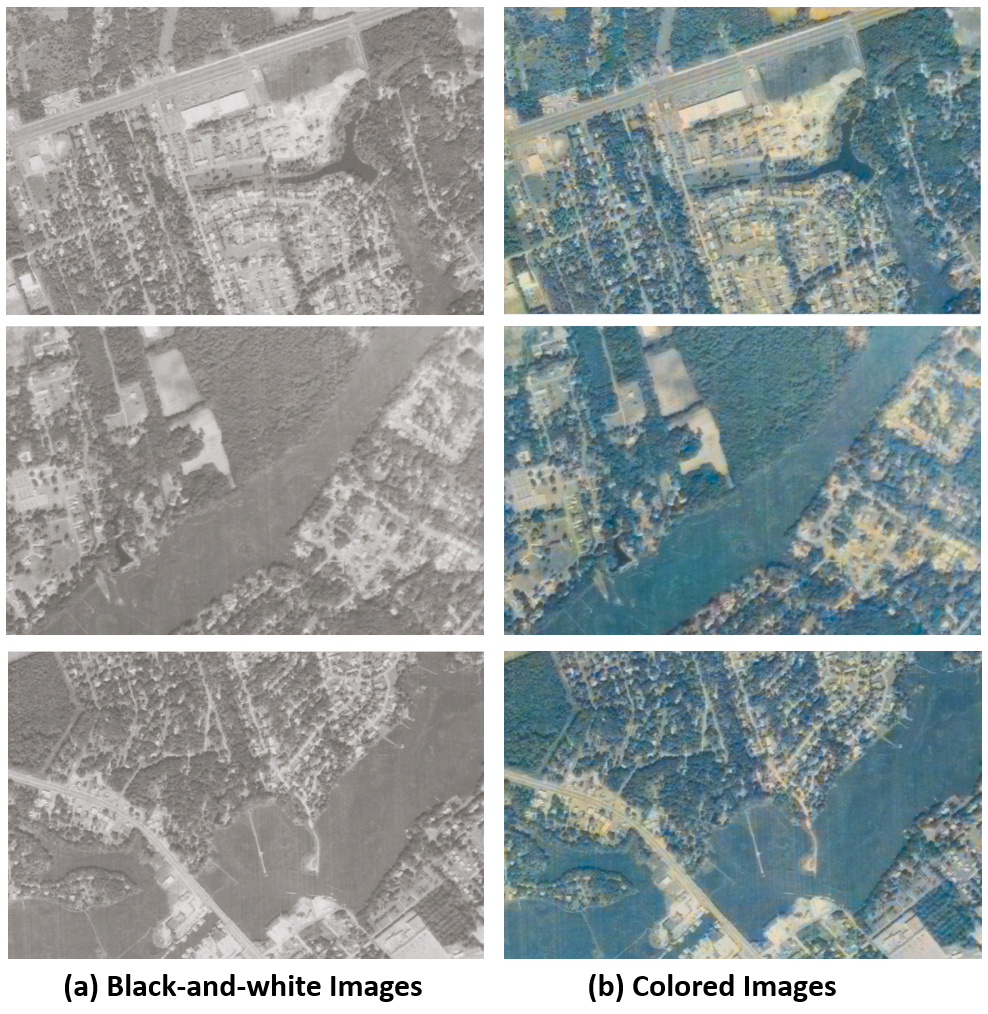}
    \caption{Comparison between the original black-and-white aerial image (left) and its colored output (right).\\
    Source: Images extracted from archived photographs \citep{usc_library_south_2024} and enhanced by authors.}
    \label{fig:Colorization}
\end{figure}

Figure~\ref{fig:Colorization} showcases the transformation of a sample grayscale aerial image into a colored version. The colorization process enhances visual interpretability by assigning plausible hues to rooftops, roads, and vegetation areas. These more realistic color tones facilitate the differentiation of building structures from their surroundings, thereby improving downstream detection performance.

\begin{figure}[H]
    \centering
    \includegraphics[width=1\linewidth]{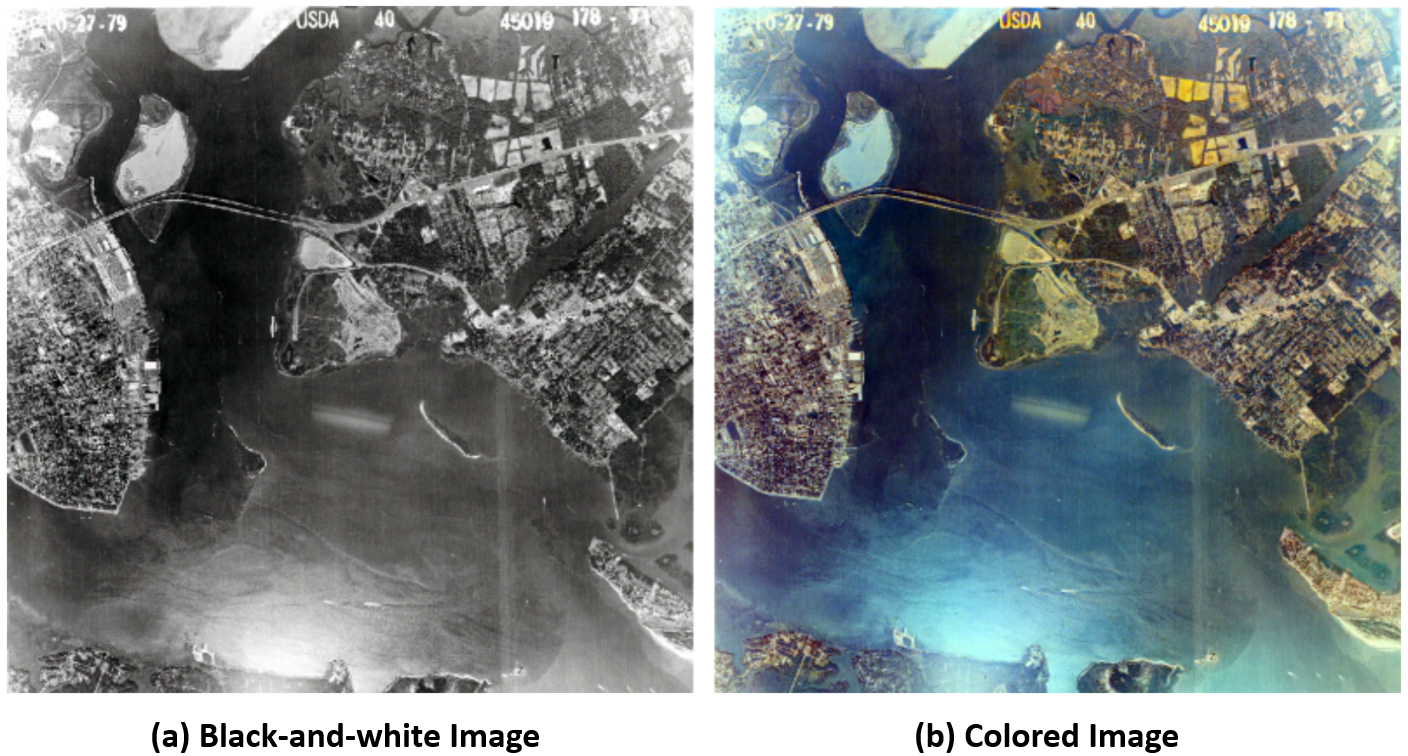}
    \caption{Overall colorization result of a complete historical aerial image tile from Charleston.\\
    Source: Images extracted from archived photographs \citep{usc_library_south_2024} and enhanced by authors.}
    \label{fig:overall_colorization}
\end{figure}

Furthermore, Figure~\ref{fig:overall_colorization} presents the colorization of an entire historical aerial image tile from Charleston. This large-scale visualization underscores how colorization effectively differentiates urban features across extensive geographic areas.

\subsection{Super-resolution imaging results}

\begin{figure}[H]
    \centering
    \includegraphics[width=1\linewidth]{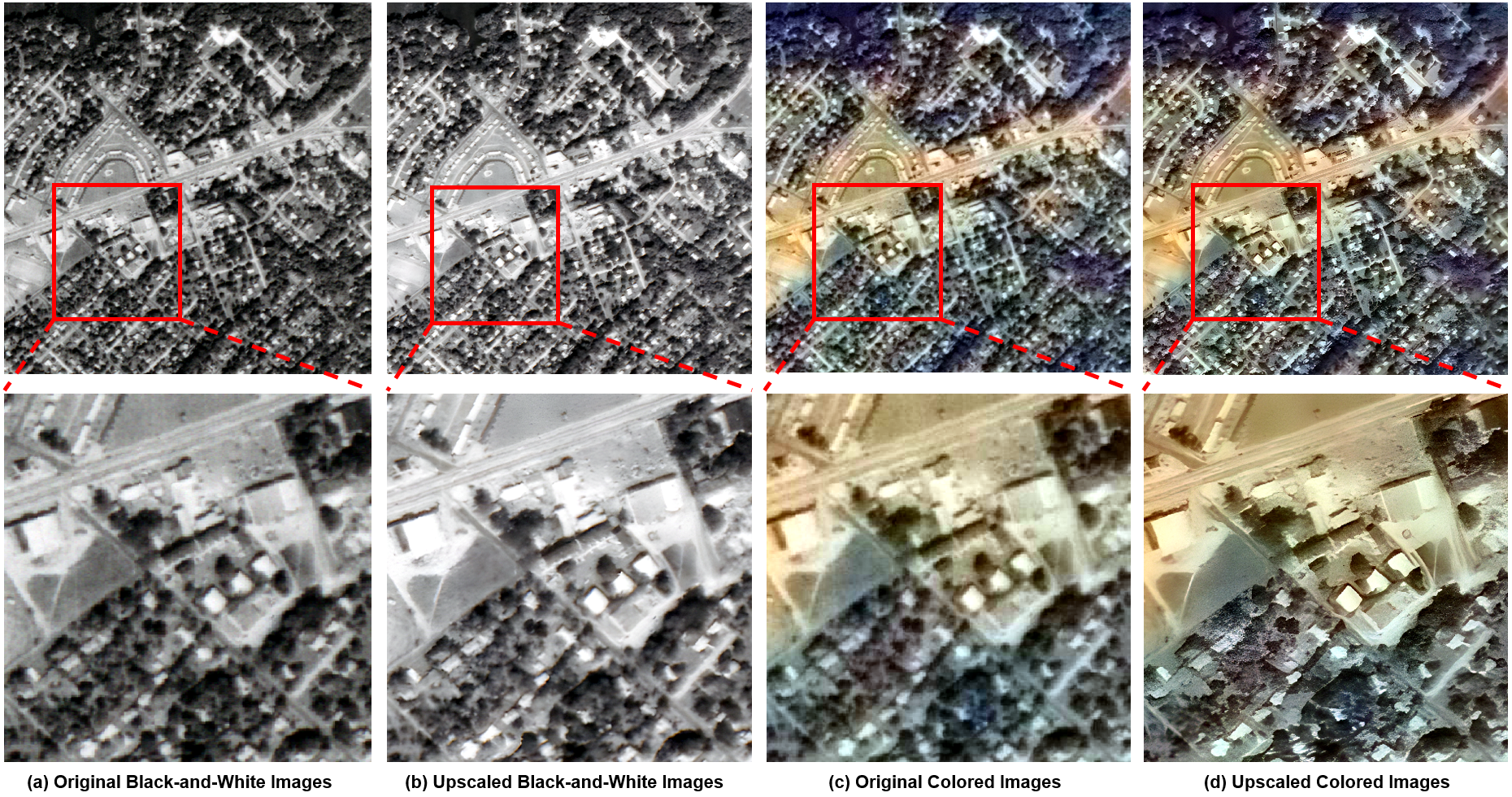}
    \caption{Super resolution results for black-and-white and colored aerial images. Left: original images; right: super-resolved outputs.\\
    Source: Images extracted from archived photographs \citep{usc_library_south_2024} and enhanced by authors.}
    \label{fig:upscale_results}
\end{figure}

We applied super resolution to both colored and black-and-white aerial images. Figure~\ref{fig:upscale_results} presents a comparison between the original and upscaled images. The left column shows the original images, while the right column displays the corresponding super-resolved outputs.

By comparing the original images to their super-resolved counterparts, we observed sharper edges, reduced noise, and enhanced textures, all of which are essential for accurately detecting rooftops. These improvements in image quality contribute to better performance in object detection tasks, potentially leading to higher mean Average Precision (mAP) and recall when training models on historical aerial imagery.

\subsection{Object detection models training results}

We experimented with four variations of the dataset: (1) original black-and-white images; (2) original color images; (3) upscaled, black-and-white images; and (4) upscaled, color images.

Our results indicate that models trained on the original data achieve mean Average Precision (mAP) values below 50\%, whereas those trained on upscaled (zoomed-in) data exceed 60\%. This underscores the importance of image super-resolution in enhancing detection performance. Therefore, we focused our comparative analysis on the upscaled data.

\begin{figure}[H]
    \centering
    \includegraphics[width=0.75\linewidth]{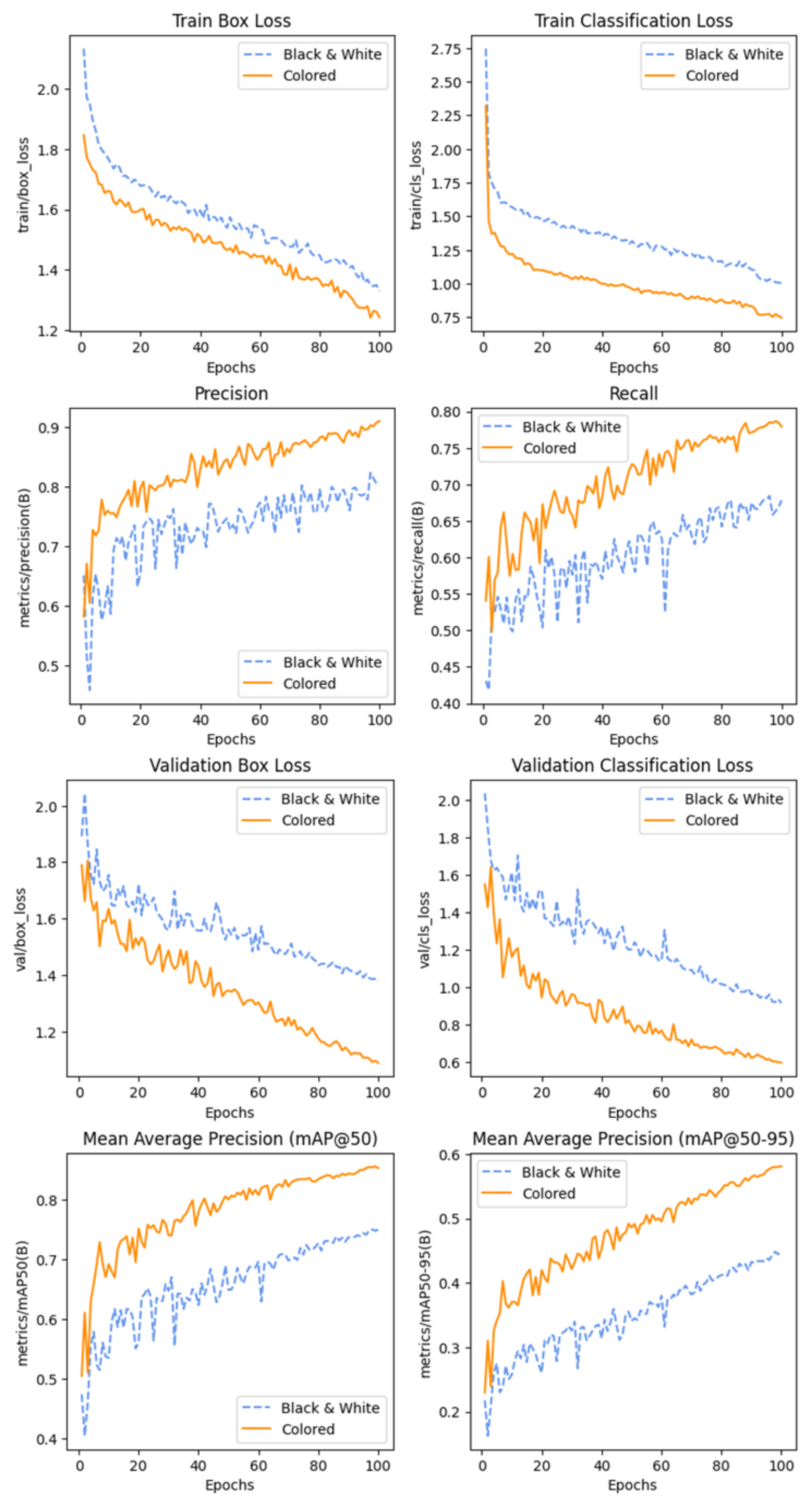}
    \caption{YOLOv11 training results comparing the upscaled black-and-white and upscaled colored datasets. The colored dataset achieves higher mAP and recall, underscoring the utility of colorization for historical rooftop detection.\\
    Source: Graph generated by authors through Python 3.10.}
    \label{fig:training_results}
\end{figure}

Figure~\ref{fig:training_results} shows the YOLOv11 training metrics for models trained on black-and-white and colored images, both upscaled from original photographs. The model trained on colored images consistently performs better across all key metrics. It reaches a final mAP@50 of 0.852—about 10\% higher than the 0.748 from the model trained on black-and-white images. Its recall is also higher at 0.784 compared to 0.657, suggesting that color helps the model identify more true positives and reduce false negatives.

Classification loss trends further support this observation. The color-trained model converges more rapidly and stably, with final training and validation losses near 0.4 and 0.6, respectively. In contrast, the model trained on black-and-white images shows higher and more fluctuating losses. The mAP@50--95 metric also favors color, with over a 10-point advantage at convergence.

Collectively, these results highlight the crucial role of colorization in improving object detection performance in historical aerial imagery. The findings suggest that training on colored data improves both classification accuracy and localization precision, ultimately leading to higher overall detection performance. This underscores the necessity of incorporating color enhancement techniques when processing historical grayscale datasets for deep-learning-based object detection.

\begin{figure}[H]
    \centering
    \includegraphics[width=\linewidth]{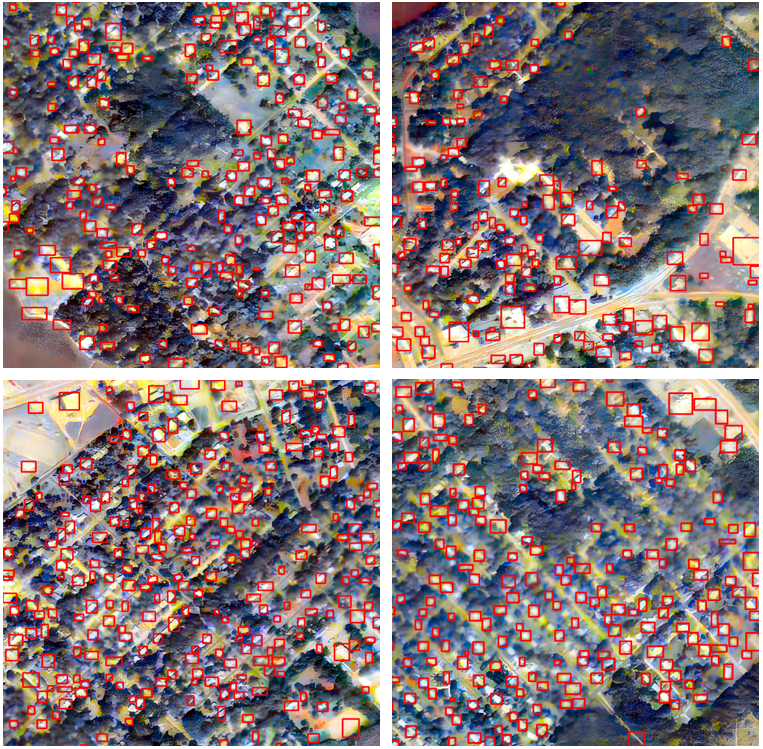}
    \caption{Test results on upscaled colored images.\\
    Source: Rooftop detection images created by authors through Python 3.10 based on images extracted from archived photographs \citep{usc_library_south_2024}.}
    \label{fig:test-result}
\end{figure}

Figure~\ref{fig:test-result} presents representative examples of rooftop detection results generated by YOLOv11n, trained on colored and super-resolved historical aerial imagery. While the model generally performs robustly, certain limitations remain evident. Visual inspection reveals two primary issues: (1) some visible rooftops were not detected, suggesting difficulty in identifying rooftops with low contrast or indistinct features, and (2) several road segments were incorrectly classified as rooftops, likely due to visual similarities in color or texture. These errors underscore the need for further refinement, particularly in improving the model’s ability to differentiate rooftops from other structurally similar urban features. A more detailed discussion is provided in Section~\ref{sec:error_analysis}.

To quantitatively evaluate the benefits of our two-stage enhancement pipeline, we systematically trained and compared three advanced object detection models—YOLOv11, Faster R-CNN, and DETR—on enhanced datasets (colored versus original black-and-white imagery). Table~\ref{tab:three_model_results} summarizes the performance of these models using mean Average Precision (mAP@50) and recall metrics.

Across all tested models, YOLOv11 consistently achieved the highest performance, especially when trained on the enhanced colored data, attaining an mAP of 85.2\% and a recall of 78.4\%. This represents significant improvements—approximately 10\% higher in absolute terms compared to training on upscaled black-and-white images. Although Faster R-CNN and DETR had lower overall performance than YOLOv11, they similarly benefited from the colorization and super-resolution preprocessing, demonstrating improvements in mAP and recall of around 5–7\%.

\begin{table}[htbp]
    \centering
    \caption{Comparison of YOLOv11, Faster R-CNN, and DETR on upscaled black-and-white vs.\ color images.}
    \label{tab:three_model_results}
    \begin{tabularx}{\textwidth}{l *{4}{>{\centering\arraybackslash}X}}
    \hline
    \textbf{Model} & \textbf{mAP@50 (B\&W)} & \textbf{Recall (B\&W)} & \textbf{mAP@50 (Colored)} & \textbf{Recall (Colored)} \\
    \hline
    \textbf{YOLOv11} & 0.7499 & 0.6784 & 0.852 & 0.784 \\
    \textbf{Faster R-CNN} & 0.622 & 0.512 & 0.673 & 0.563 \\
    \textbf{DETR} & 0.613 & 0.550 & 0.682 & 0.690 \\
    \hline
    \end{tabularx}
\end{table}

These results underscore the value of colorization and super-resolution as preprocessing steps, which markedly improve rooftop detection reliability across different model architectures. The observed errors also suggest directions for further refinement, such as incorporating additional contextual or texture-based features to distinguish rooftops from visually similar urban structures.

\section{Discussion}

\subsection{Implications for historical GIS and remote sensing}
\label{sec:error_analysis}

This study highlights the potential of deep learning methods to extract valuable insights from historical aerial imagery. By addressing challenges related to common limitations of low resolution and high noises, the presented workflow demonstrates that advanced detection models can be effectively adapted to historical contexts through appropriate preprocessing and input guidance. Our experiments contribute to the growing literature in historical GIS, offering versatile tools and methods applicable across disciplines, including urban studies and environmental history.

Furthermore, our work aligns with current trends in remote sensing and GIS that integrate historical data with modern analytical techniques and significantly augments the potential and reliability of using historical remote sensing data \citep{ratajczak_automatic_2019, kindermann_combining_2023}. Such integration enables comprehensive analyses of long-term urban and ecological changes, thereby enhancing policy-making and planning based on historical trends.

Ultimately, the selection of a detection method entails a trade-off among precision, scalability, and resource requirements. Researchers and practitioners should carefully balance these factors to tailor their workflows to the specific needs of their projects.

\subsection{The influence of image enhancement}
\label{subsec:influence_image_enhancement}

A key finding of this study is that image enhancement through colorization and super-resolution significantly improves rooftop detection performance by augmenting both the spectral and spatial fidelity of historical aerial imagery. Colorization provides critical spectral cues that help distinguish building rooftops from other urban features such as roads, parking lots, and vegetation. In many historical aerial images, subtle differences in texture or grayscale intensity are insufficient for deep-learning models to accurately segment or classify rooftops. By adding plausible color information, colorization effectively bridges the domain gap between archival black-and-white photographs and the modern RGB imagery on which most detection architectures are trained.

Super resolution further contributes by sharpening edges, reducing noise, and restoring details that might otherwise be lost in low-resolution scans. This is particularly valuable for delineating complex building outlines and detecting smaller structures that would remain indistinguishable at coarser scales. As shown in our experiments, models trained on upscaled colored images consistently outperformed those trained on their black-and-white or lower-resolution counterparts, achieving a mean Average Precision (mAP) of up to 85\%.

Nevertheless, image enhancement methods integrated in our experiments also have a few potential limitations. Colorization may produce artifacts if the underlying grayscale patterns are misinterpreted by the generative model, leading to unrealistic hues or “bleeding” of color across object boundaries \citep{shafiq_image_2023}. Similarly, super resolution can “hallucinate” details that do not exist in the original data, potentially biasing detection outcomes in borderline cases\citep{greza_gan-based_2024}. Despite these caveats, our results suggest that the benefits of image enhancement outweigh the risks in most scenarios, particularly when dealing with historically valuable datasets that lack sufficient resolution or color channels.

\subsection{Error analysis and failure cases}
As part of the interpretative analysis of our findings, we examined persistent errors encountered during rooftop detection, identifying potential causes and outlining specific opportunities for future improvements:

\begin{enumerate}
    \item Missed Rooftops: Some rooftops remained undetected, particularly in degraded grayscale images with extreme overexposure or noise. In such cases, the colorization model struggled to recover subtle luminance variations, producing flat regions misinterpreted as background. Enhancing the model’s robustness in handling extreme degradation could mitigate this issue.
    \item False Positives (Road Misclassification): The model occasionally misclassified elongated road segments as rooftops. This likely stems from artifacts introduced by the super-resolution process, where enhanced edges create high-frequency details resembling rooftop textures. Future improvements may involve integrating multi-scale features or additional semantic context during training.
    \item Colorization Artifacts: Color bleeding and inconsistent hue mapping, particularly along building edges, sometimes led to false detections. Refining the DeOldify model with improved perceptual loss components or attention mechanisms could minimize these distortions.
    \item Super-Resolution Hallucination: The Real-ESRGAN model sometimes introduced artificial textures, improving visual clarity but occasionally misleading the detection algorithm. A comparative analysis with high-resolution ground truth images could help assess the reliability of these enhancements.
\end{enumerate}

Although the proposed pipeline greatly improves detection performance, in the future by addressing these failures with targeted improvements and ablation studies, we may be able to further improve the robustness of roof detection in historical aerial imagery.

\subsection{Limitations of large-scale segmentation models}

As a preliminary exploration to assess feasibility, we experimented briefly with large-scale segmentation models (SAM and SAM2) in a zero-shot and prompt-based context without additional fine-tuning \citep{sultan_geosam_2024, ravi_sam_2024,kirillov_segment_2023}. Unfortunately, our results indicate that these models underperformed in the specific task of historical rooftop detection. First, outputs from both models consistently omitted many clearly visible rooftops, resulting in high false-negative rates. This limitation likely stems from the significant domain gap between the contemporary imagery used to train these models and the historical aerial photographs employed in our study.

\begin{figure}[H]
    \centering
    \includegraphics[width=1\linewidth]{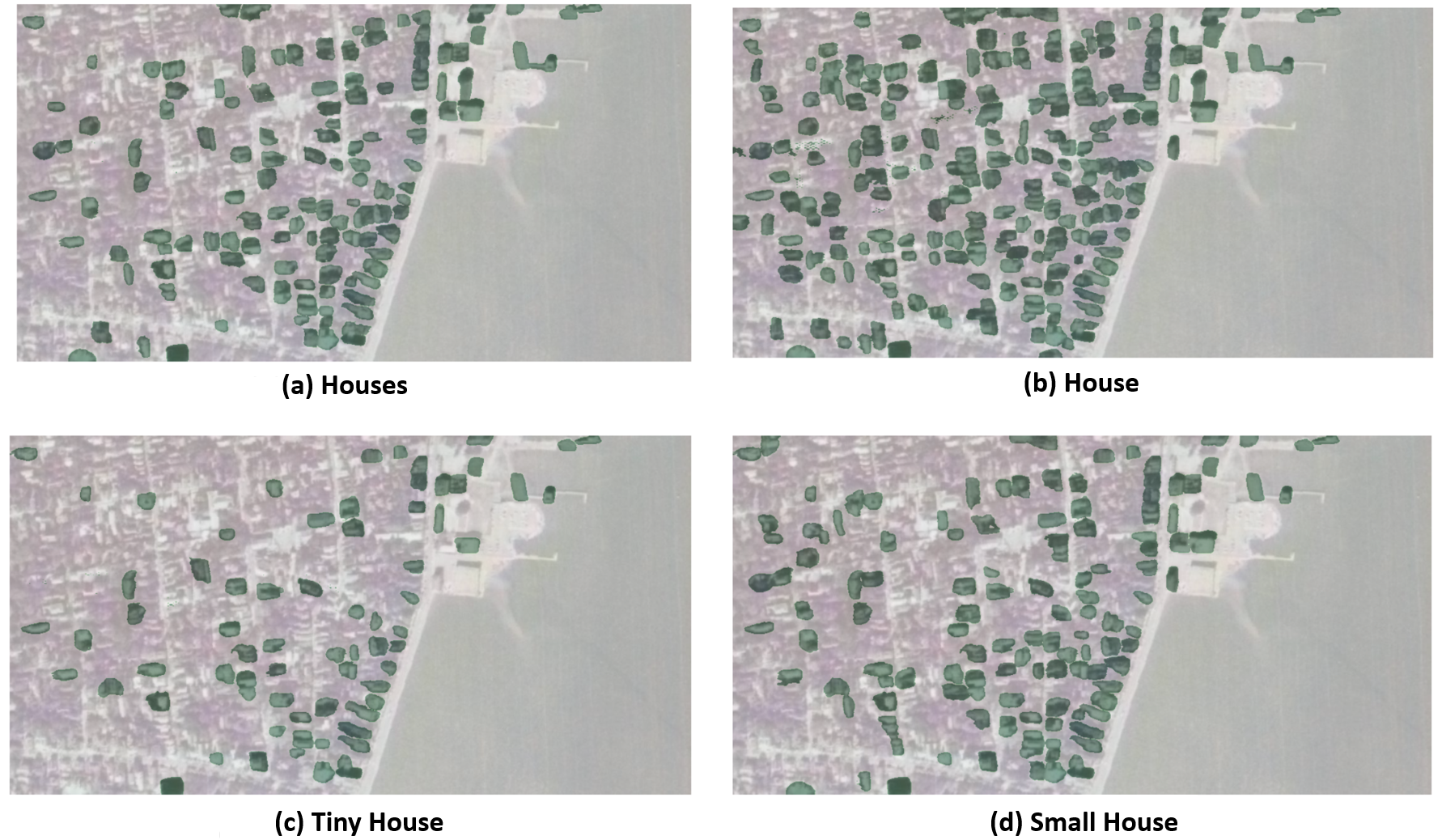}
    \caption{Results of text-prompt learning with SAM2, tested with four different prompts.\\
    Source: testing images created by authors through Python 3.10 based on images extracted from archived photographs \citep{usc_library_south_2024}.}
    \label{fig:text_prompt}
\end{figure}

Second, when employing text prompts with SAM2, the outputs proved overly sensitive to textual variations. As illustrated in Figure~\ref{fig:text_prompt}, slight differences in prompts led to substantial changes in segmentation outputs, reducing consistency and reliability.

These observations suggest that despite the general effectiveness of large-scale segmentation models in other domains, their direct zero-shot application is insufficient for historical rooftop detection without further adaptation or domain-specific training.

\subsection{Rationale for excluding diffusion models in colorization}

Recent advancements in diffusion models have demonstrated impressive capabilities in general image synthesis and image-to-image translation tasks, including colorization \citep{ho_denoising_2020, saharia_palette_2022}. Despite their visual realism, we opted not to employ diffusion-based methods primarily because diffusion models inherently lack explicit constraints to preserve geometric or spatial fidelity. Their outputs frequently introduce minor yet consequential distortions to landforms, boundaries, or built structures, presenting unacceptable risks for downstream geospatial tasks such as rooftop detection or urban morphological analysis \citep{rombach_high-resolution_2022, saharia_palette_2022}. These spatial inconsistencies compromise both the accuracy of subsequent detection models and the interpretability of temporal urban change analyses.

Additionally, diffusion models typically require extensive computational resources, including significant GPU memory, extended training periods, and large-scale, high-quality datasets to achieve stable results. These demands are often difficult to satisfy within the context of historical aerial datasets, which frequently exhibit considerable variability in resolution, contrast, and overall scanning quality.

Lastly, diffusion-based approaches can be complex to fine-tune and integrate into an end-to-end processing pipeline \citep{ruiz_dreambooth_2023}. In contrast, our chosen GAN-based method offers a balanced trade-off among visual quality, computational efficiency, and ease of integration, facilitating iterative experimentation and enabling scalable processing of archival aerial imagery with minimal geometric distortion.

\section{Conclusion}
In this study, we introduce a GAN-enhanced deep learning framework specifically designed for rooftop detection from historical aerial imagery. By integrating two critical preprocessing steps—image colorization using DeOldify and super-resolution via Real-ESRGAN—our framework effectively addressed the inherent limitations of historical black-and-white aerial photographs, such as low spatial resolution, grayscale-only channels, and archival noise. Through extensive experimentation using three state-of-the-art object detection architectures (YOLOv11, Faster R-CNN, and DETR), we found consistent evidence that the addition of color information significantly improves detection performance. Notably, our quantitative evaluation revealed that YOLOv11 achieved a mean Average Precision (mAP) of 0.852 and a recall of 0.784 when trained on colored and super-resolved images, outperforming the same model trained on black-and-white images by approximately 10\% in mAP and recall. Further, visual inspections confirmed that colorization particularly enhanced boundary clarity and object distinction, enabling more precise rooftop identification and reducing confusion with roads and similar urban structures. However, certain challenges remained, including occasional missed detections of less distinct rooftops and false positives due to structural similarities between rooftops and roads. Overall, these detailed findings underscore the importance and effectiveness of generative image enhancement techniques as a necessary preprocessing step, significantly advancing the robustness of deep-learning-based rooftop detection from historical aerial data. The proposed methodology not only serves as practical guidance for remote-sensing practitioners utilizing historical archives but also highlights the broader potential for incorporating generative techniques into other object detection scenarios facing similarly challenging imaging conditions.

\bibliographystyle{apalike} 
\bibliography{references}

\end{document}